\pdfoutput=1

\documentclass[11pt]{article}

\usepackage[final]{acl}

\usepackage{times}
\usepackage{booktabs}
\usepackage{authblk}
\usepackage{latexsym}
\usepackage{multirow}
\usepackage{amssymb}
\usepackage{float}
\usepackage{graphicx}
\usepackage{amsmath}
\usepackage{algorithm}
\usepackage[noend]{algpseudocode}
\usepackage[T1]{fontenc}

\usepackage[utf8]{inputenc}
\usepackage{microtype}

\usepackage{inconsolata}

\usepackage{subfigure}
\usepackage{placeins}
\title{\includegraphics[scale=0.1]{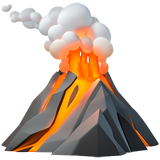} {\Ours}: Mitigating Multimodal Hallucination through \\ Self-Feedback Guided Revision}

\makeatletter
\renewcommand\AB@affilsepx{\hspace{0.5cm} \protect\Affilfont}
\makeatother

\author[1]{\textbf{Seongyun Lee}}
\author[1]{\textbf{Sue Hyun Park}}
\author[2]{\textbf{Yongrae Jo}}
\author[1]{\textbf{Minjoon Seo}}
\affil[1]{KAIST AI}
\affil[2]{LG AI Research \protect \\[3ex] \texttt{\{seongyun, suehyunpark, minjoon\}@kaist.ac.kr yongrae.jo@lgresearch.ai}}

\newcommand{\Ours}{\textsc{Volcano}}

\begin{document}
\maketitle

\begin{abstract}
Large multimodal models suffer from multimodal hallucination, where they provide incorrect responses misaligned with the given visual information. Recent works have conjectured that one of the reasons behind multimodal hallucination is due to the vision encoder failing to ground on the image properly. To mitigate this issue, we propose a novel approach that leverages self-feedback as visual cues. Building on this approach, we introduce \textbf{{\Ours}}, a multimodal self-feedback guided revision model. {\Ours} generates natural language feedback to its initial response based on the provided visual information and utilizes this feedback to self-revise its initial response. {\Ours} effectively reduces multimodal hallucination and achieves state-of-the-art on MMHal-Bench, POPE, and GAVIE. It also improves on general multimodal abilities and outperforms previous models on MM-Vet and MMBench. Through qualitative analysis, we show that {\Ours}'s feedback is properly grounded on the image than the initial response. This indicates that {\Ours} can provide itself with richer visual information through feedback generation, leading to self-correct hallucinations. We publicly release our model, data, and code at \href{https://github.com/kaistAI/Volcano}{github.com/kaistAI/Volcano}.
\end{abstract}

\section{Introduction}
\label{sec:intro}
Recent large multimodal models (LMMs) use substantial image-text or video-text pairs to help instruct-tuned large language models (LLMs) comprehend visual features produced by vision encoders  \citep{alayrac2022flamingo, liu2023improved, liu2023visual, chen2023shikra, peng2023kosmos2, dai2023instructblip, zhu2023minigpt4, ye2023mplugowl, li2023otter, zhang2023llamaadapter, su2023pandagpt, maaz2023videochatgpt}. With the introduction of fine-tuning methods such as visual instruction tuning \citep{liu2023improved, liu2023visual}, LMMs are now evolving into assistants capable of understanding the world through multiple channels, akin to humans.

\begin{figure}[t!]
\includegraphics[width=1.0\linewidth]{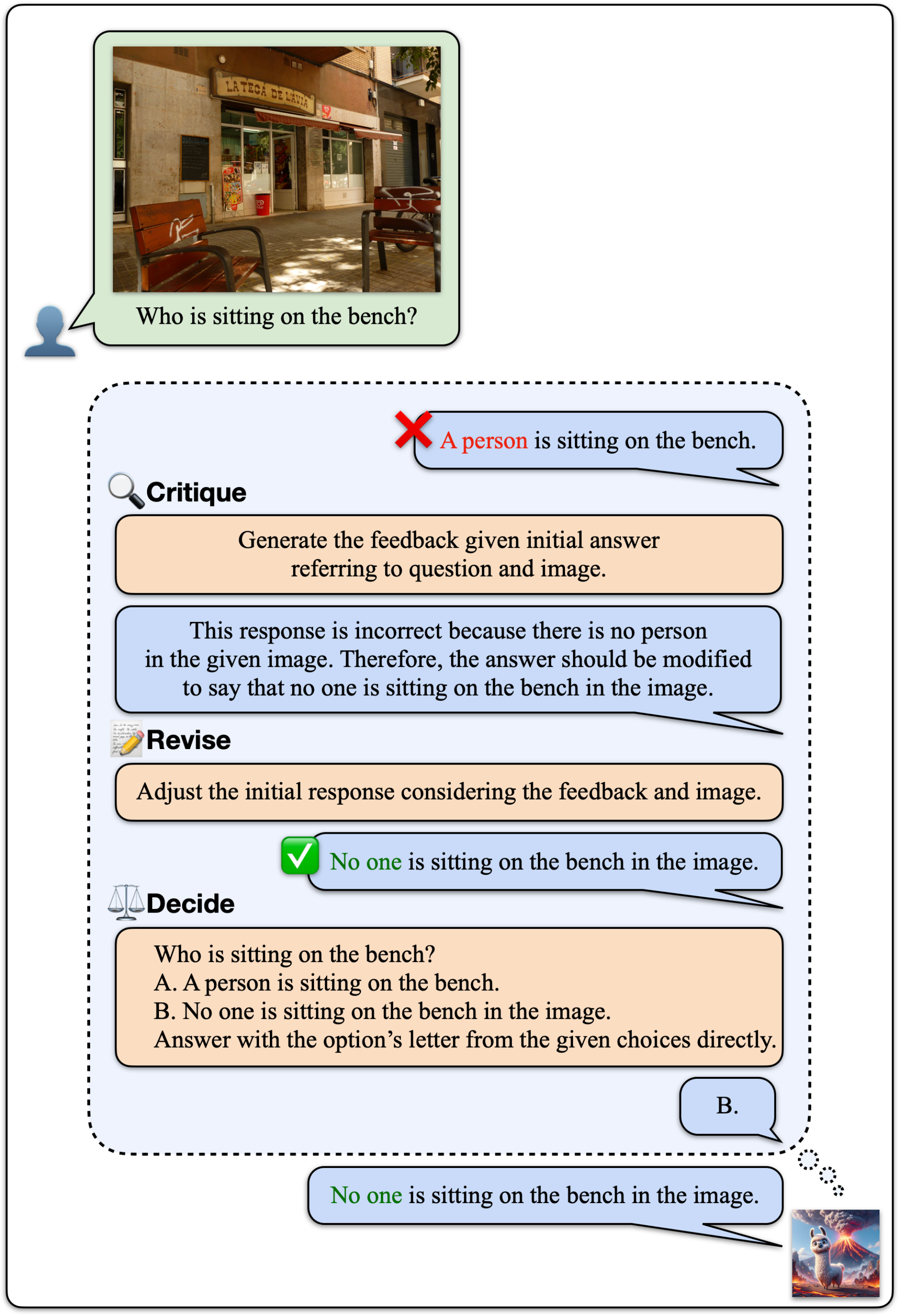}
\caption{Overview of {\Ours}. This example illustrates the process undertaken by {\Ours} for a question in the MMHal-Bench dataset. Before giving the response, {\Ours} goes through a \textit{critique-revise-decide} process. It critiques its initial response with natural language feedback, revises the response based on the feedback, and decides whether to accept the revised answer.}
\label{fig:figure1}
\end{figure}

Despite the impressive performance observed on various benchmark tasks and qualitative outcomes, these models grapple with an issue called \textit{multimodal hallucination}, where they produce responses that do not align with the visual information given in the question.
Recent work \citep{zhai2023halleswitch} demonstrates that multimodal hallucinations can occur when the vision encoder fails to ground images accurately. In other words, LMMs tend to rely more on their parametric knowledge than on provided visual features, causing them to guess and generate hallucinations. \citet{wang2023evaluation} empirically shows that models attend to the previous tokens more than image features as they generate tokens misaligned with the given image.

In this paper, we propose a novel method that utilizes natural language feedback to enable the model to correct hallucinated responses by providing detailed visual information. Building on this method, we introduce \textbf{{\Ours}}\footnote{We call our model {\Ours} because it frequently erupts \textit{LLaVA}}, a multimodal self-feedback guided revision model. {\Ours} is trained to first generate an initial response based on the given image and question, then sequentially revises the response until it determines that no more improvement is required. We collect our training data for multimodal feedback and revision using proprietary LLMs.

To verify the efficacy of {\Ours} in reducing multimodal hallucination, we evaluate its performance on multimodal hallucination benchmarks \citep{sun2023aligning, li2023evaluating, liu2023mitigating}. Results demonstrate consistent performance improvements across all benchmarks. Notably, when compared to previous methods specialized in mitigating multimodal hallucination \citep{zhou2023analyzing, sun2023aligning, yin2023woodpecker}, {\Ours} showcases a 24.9\% enhancement, underscoring its effectiveness in addressing the challenge. Further, on multimodal understanding benchmarks \citep{liu2023mmbench, yu2023mmvet}, it is also shown effective in understanding and reasoning about visual concepts.

Through qualitative analysis, we find that the generated feedback attends to the image with higher intensity and higher coverage of features in the image. These findings explain that feedback carries fine-grained visual information. Even if the vision encoder fails to properly ground, the feedback can still guide the LLM to improve upon the hallucinated response, supporting the role of feedback in our proposed method.

Our contributions are summarized as follows:
\begin{description}
\item 1. We introduce {\Ours}, a self-feedback guided revision model that effectively mitigates multimodal hallucination. It achieves state-of-the-art performance on multimodal hallucination benchmarks and multimodal understanding benchmarks.
\item 2. Our qualitative analysis shows that {\Ours}'s feedback is rooted in the image, conveying rich visual details. This illustrates that feedback can offer guidance to reduce multimodal hallucination, even if the vision encoder imprecisely encodes the image and the model misinterprets the image initially.
\item 3. We open-source {\Ours} (7B \& 13B), along with data and code for training and inference.
\end{description}

\section{Related work}
\label{sec:related}
\subsection{Multimodal hallucination} 
\label{subsec:hal}
Unlike language hallucination, where fabrication of unverifiable information is common \citep{10.1145/3571730,zhang2023sirens, li2023halueval}, multimodal hallucination typically involves verifiable information misaligned with the input visual content. This phenomenon has been predominantly explored in the context of object hallucination, where generated content includes objects that are inconsistent with or absent from the target image \citep{rohrbach-etal-2018-object, 9706727, li2023evaluating, liu2023mitigating, zhai2023halleswitch}. More complex forms of multimodal hallucination, such as holistic misrepresentations involving entire scenes or environments, have only begun to be recognized and documented in recent studies \citep{sun2023aligning}. 

To uncover the cause of failure in grounding, previous works analyze either the visual or language side. \citet{zhai2023halleswitch} pinpoints the lack of preciseness in visual features produced by the vision encoder. Other studies \citep{li2023evaluating, liu2023mitigating, wang2023evaluation} focus on the tendency of LLMs to generate words more in line with common language patterns rather than the actual visual content. The error may be further exacerbated by autoregressive text generation \citep{rohrbach-etal-2018-object, zhang2023language, zhou2023analyzing}.

\subsection{Self-correcting from feedback}
\label{subsec:feedback}
Learning from feedback can align LLMs to desired outcomes, to better follow instructions via human preference feedback \citep{ouyang2022training}, preference feedback generated by AI itself \citep{lee2023rlaif, dubois2023alpacafarm}, or even fine-grained feedback \citep{wu2023finegrained, lightman2023lets}. Compared to preference and fine-grained feedback which provide scalar values as training signals, natural language feedback provides more information \citep{scheurer2022training, ma2023eureka} and has been effective for language models to correct outputs, especially for \textit{self-correction} \citep{welleck2022generating, pan2023automatically}. Inspired by successful iterative self-refining language models \citep{madaan2023selfrefine, selfee2023, shinn2023reflexion, gou2024critic}, to the best of our knowledge, we are the first to achieve improvement in multimodal models through self-feedback guided refinement.

\subsection{Mitigating multimodal hallucination} 
\label{subsec:mitigating}
Previous methods for mitigating multimodal hallucinations have varied in their focus, including enhancing the quality of instruction tuning data,  model training methodologies, and implementing post-hoc refinements. LRV-Instruction dataset \citep{liu2023mitigating} ensures the balance of both negative and positive instructions and VIGC \citep{wang2023vigc} iteratively generates and corrects instructions to reduce hallucinated samples in training data. Adapting reinforcement learning from human feedback (RLHF) to train a single reward model as in LLaVA-RLHF \citep{sun2023aligning} or training multiple or even without no reward models as in FDPO \citep{gunjal2023detecting} has proven effective as well. LURE \citep{zhou2023analyzing} trains a revision model to detect and correct hallucinated objects in the base model's response. Woodpecker \citep{yin2023woodpecker} breaks down the revision process into multiple subtasks where three pre-trained models apart from the base LMM are employed for the subtasks.

Unlike models using reinforcement learning, our approach does not require reward model training. Also, contrary to revision-only methods, our method trains a model to \textit{self}-revise, eliminating the need for extra modules. Furthermore, we introduce natural language feedback before the revision process. This feedback serves a dual purpose: it revisits the visual features for enhanced clarity and specifically pinpoints the hallucinated elements that require correction, thereby enriching the information available for more effective revision.

\section{{\Ours}}
\label{sec:volcano}

\begin{algorithm}[t]
\caption{Feedback guided self-revision}
\begin{algorithmic}[1]
\State \textbf{Input:} \textit{model $M$}, \textit{image $I$}, \textit{question $Q$}
\State \textit{$R_{initial} = M(I, Q)$}
\State \textit{$R_{best} = R_{initial}$}
\For{up to 3 iterations}
    \State \textit{$F = M(I, Q, R_{best})$} \hspace{0.3cm}
    \State \textit{$R_{revised} = M(I, Q, R_{best}, F)$} \hspace{0.3cm}
    \State \textit{$R_{decided} = M(I, Q, R_{best}, R_{revised})$} \hspace{0.3cm}
    \If {$R_{decided} == R_{best}$}
        \State \textbf{break}
    \Else
        \State \textit{$R_{best} = R_{revised}$}
    \EndIf
\EndFor
\State \textbf{return} \textit{$R_{best}$}
\end{algorithmic}
\label{alg:algorithm}
\end{algorithm}

\begin{figure*}[ht]
\includegraphics[width=1.0\linewidth]{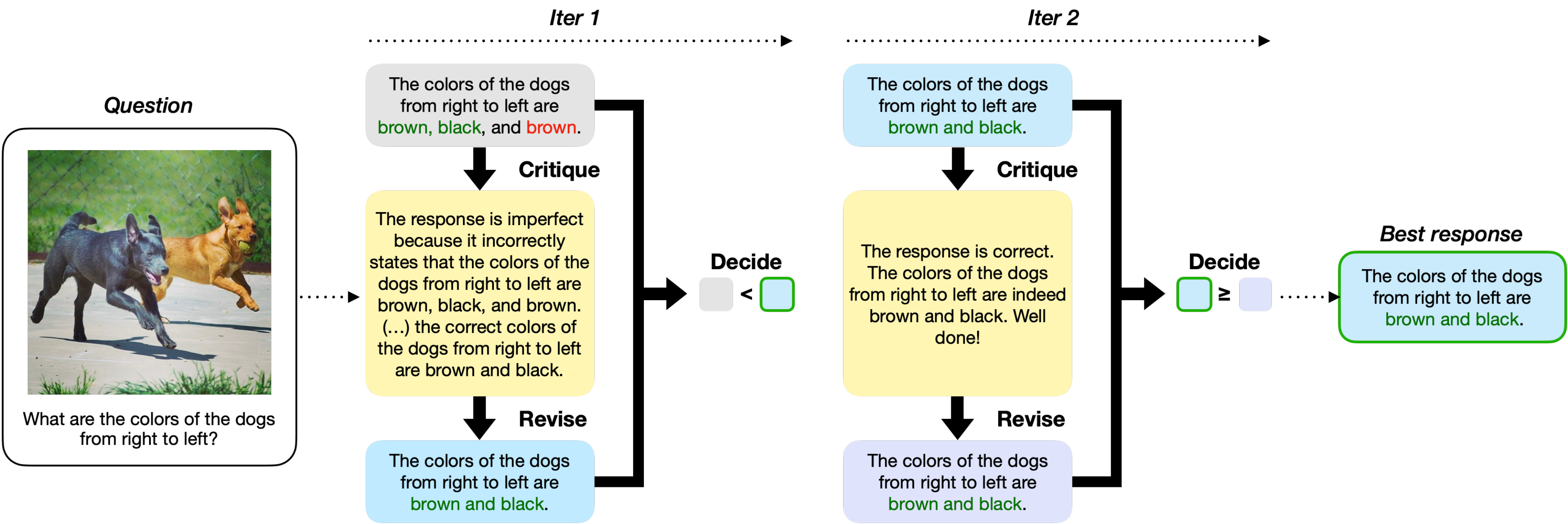}
\caption{Overall process of {\Ours}. {\Ours} is a multimodal self-feedback guided revision model that takes an image and a question and then generates an improved response based on the self-feedback.}
\label{fig:figure2}
\end{figure*}

{\Ours} is a single LMM to generate initial responses, feedback, and revisions, as well as decisions to accept revisions. It follows a sequential procedure of an iterative critique-revision-decide loop. In Section~\ref{subsec:iterative}, we introduce the process by which {\Ours} self-revises its responses iteratively. Section~\ref{subsec:data} describes the collection of multimodal feedback and revision data used to train {\Ours}. Finally, Section~\ref{subsec:imple} provides detailed information about the models and data used in our study. The overall process is explained in Algorithm~\ref{alg:algorithm} and illustrated in Figure~\ref{fig:figure2}.

\subsection{Iterative self-revision}
\label{subsec:iterative}
{\Ours} is a single model that generates improved responses through a sequential process of four stages. First, similar to other LMMs, it generates an initial response $R_{initial}$ for the image $I$ and question $Q$ and initializes the best response $R_{best}$ with $R_{initial}$. This stage is performed only once in the process of creating the final response. Second, it generates feedback $F$ based on the $R_{best}$ (\textbf{stage 1}). Using this feedback, it self-revises the $R_{best}$ (\textbf{stage 2}). Since there is no guarantee that the revised response $R_{revised}$ will be better than the existing $R_{best}$, there is a need to determine which response is better for the given $Q$ and $I$. At this point, {\Ours} is given $Q$, $I$, and both responses, and it goes through the process of deciding which response is better (\textbf{stage 3}). The order of $R_{revised}$ and $R_{best}$ in stage 3 is randomized to prevent the positions from affecting the results \citep{wang2023large}. If the model decides that $R_{revised}$ is better than $R_{best}$, then $R_{best}$ is updated with $R_{revised}$ and the procedure from stage 1 to stage 3 is repeated, with the predetermined maximum number of iterations. Otherwise, the loop is early-stopped, and $R_{best}$ is selected as the final output. The prompts for inference at each stage are in Appendix~\ref{subsec:inference_prompts}.

\begin{figure}[h!]
\includegraphics[width=1.0\linewidth]{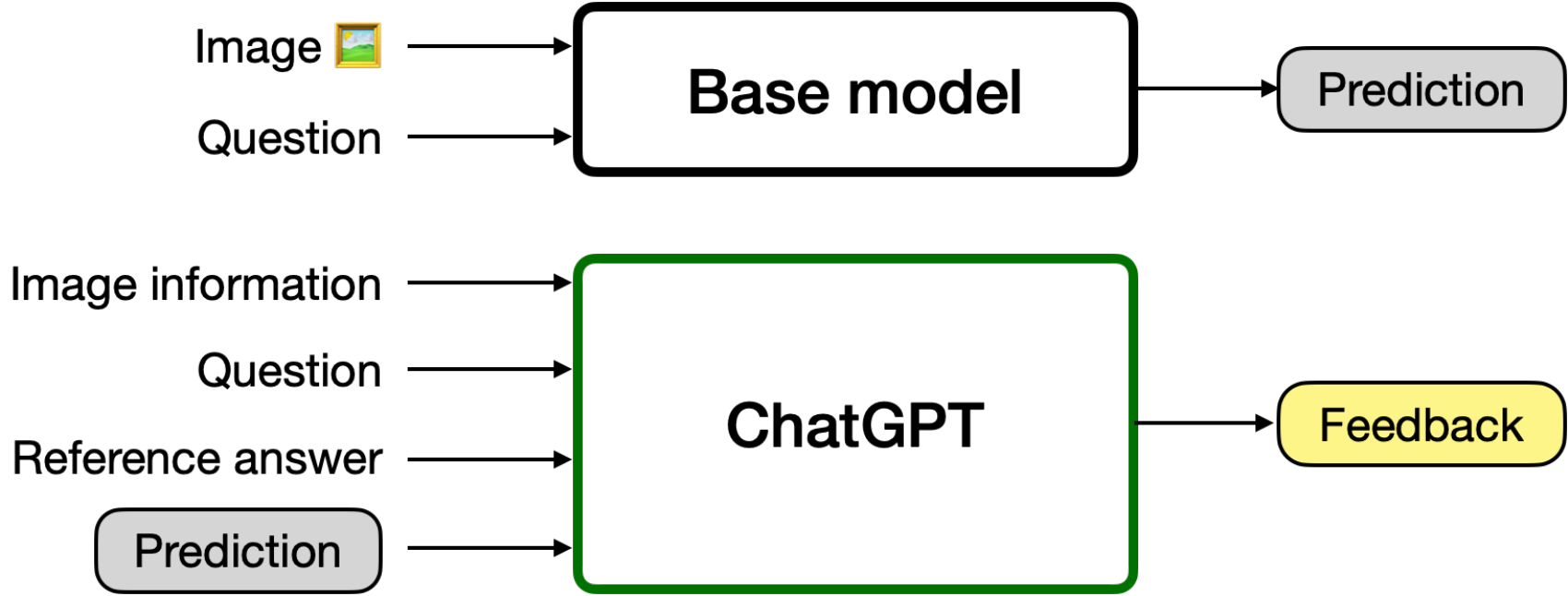}
\caption{Data collection.}
\label{fig:figure3}
\end{figure}

\subsection{Data collection}
\label{subsec:data}
To train {\Ours}, we collect initial responses for visual questions from an open-source LMM and generate feedback and revisions using a proprietary LLM as shown in Figure~\ref{fig:figure3} \citep{akyürek2023rl4f, madaan2023selfrefine, selfee2023, wang2023shepherd, kim2023prometheus}.

Since current proprietary LLMs cannot process images, we provide object details in text and image captions as a proxy for images. For each data instance, we feed the proprietary LLM image information consisting of object details and captions, question, initial response, and gold answer as reference answer, allowing the model to evaluate the given inputs and produce feedback.

The proprietary LLM might exploit the gold answer to generate the feedback, which can cause potential inaccuracies in feedback during inference time when it is not provided. To avoid this, we give the LLM clear prompts to focus on the text-formatted image details when generating feedback. When constructing the revision data, we set up a system to predict the existing gold answer as the output, using the feedback data, image, question, and initial response obtained from the previous steps as input, without involving any separate model generation process. The prompts for data collection are in Appendix~\ref{subsec:data_prompts}.

\begin{table*}[ht]
\centering
\small
\begin{tabular}{lcc|cc|ccc}
\toprule
\multirow{2}{*}{Model} & \multicolumn{2}{c|}{MMHal-Bench} & \multicolumn{2}{c|}{POPE} & \multicolumn{3}{c}{GAVIE} \\
& Score $\uparrow$ & Hal rate $\downarrow$ & Acc $\uparrow$ & F1 $\uparrow$ & Acc score $\uparrow$ & Rel score $\uparrow$ & Avg score $\uparrow$ \\
\midrule
MiniGPT-4 7B & - & - & 68.4 & 74.5 & 4.14 & 5.81 & 4.98 \\
mPLUG-Owl 7B & - & - & 51.3 & 67.2 & 4.84 & 6.35 & 5.6 \\
InstructBLIP 7B & 2.1 & 0.58 & 71.5 & 80.0 & 5.93 & 7.34 & 6.64 \\
LLaVA-SFT+ 7B & 1.76 & 0.67 & 81.6 & 82.7 & 5.95 & 8.16 & 7.06 \\
LLaVA-RLHF 7B & 2.05 & 0.68 & 81.8 & 81.5 & 6.01 & 8.11 & 7.06 \\
LLaVA-SFT+ 13B & 2.43 & 0.55 & 83.2 & 82.8 & 5.95 & 8.2 & 7.09 \\
LLaVA-RLHF 13B & 2.53 & 0.57 & 83.1 & 81.9 & 6.46 & 8.22 & 7.34 \\
\midrule
LLaVA-1.5 7B & 2.42 & 0.55 & 86.1 & 85.1 & 6.42 & 8.2 & 7.31 \\
LLaVA-1.5 13B & 2.54 & 0.52 & 86.2 & 85.2 & 6.8 & 8.47 & 7.64 \\
{\Ours} 7B & 2.6 & 0.49 & 88.2 & \textbf{87.7} & 6.52 & 8.4 & 7.46 \\
{\Ours} 13B & \textbf{2.64} & \textbf{0.48} & \textbf{88.3} & \textbf{87.7} & \textbf{6.94} & \textbf{8.72} & \textbf{7.83} \\
\bottomrule
\end{tabular}
\caption{Results on multimodal hallucination benchmarks.   The MMHal-Bench score is measured on a 0-5 scale. Hallucination rate (Hal rate) is measured as the proportion of scores less than 3. Additionally, GAVIE's Acc score (Accuracy score) and Rel score (Relevancy score) are measured on a 0-10 scale, with Avg score representing the average of Acc and Rel scores. Detailed evaluation results for each benchmark by question type are in Table~\ref{tab:Table6} and Table~\ref{tab:Table7}.}
\label{tab:Table1}
\end{table*}

\subsection{Implementation details}
\label{subsec:imple}
\paragraph{Data} To construct multimodal feedback and revision data, we utilize the LLaVA-SFT-127k dataset \citep{sun2023aligning}. We only use the first turn of each instance in the dataset. When fine-tuning {\Ours}, we use the llava-1.5-mix665k as the visual instruction dataset \citep{liu2023improved}.

\paragraph{Model} For the proprietary LLM, we employ OpenAI's gpt-3.5-turbo \citep{chatgpt}. We use the LLaVA-SFT+ 7B model to generate the initial response when creating feedback data and LLaVA-1.5 7B and 13B as backbone models of {\Ours} \citep{liu2023improved, liu2023visual}. Details of computation and hyperparameters used are in Appendix~\ref{sec:computation} and  Appendix~\ref{sec:hyperparameters}, respectively.

\section{Experiments}
\label{sec:exp}

\subsection{Benchmarks}
\label{subsec:bench}

\paragraph{Multimodal hallucination benchmarks} We use POPE \citep{li2023evaluating}, GAVIE \citep{liu2023mitigating}, and MMHal-Bench \citep{sun2023aligning} as our benchmarks to test multimodal hallucination mitigation performance. POPE and GAVIE are benchmarks for assessing object-level hallucinations in images. POPE comprises 9k questions asking if a specific object is present or not in an image. GAVIE is composed of 1k questions evaluating how accurately the response describes the image (accuracy) and how well the response follows instructions (relevancy) using GPT-4. MMHal-Bench aims to evaluate the overall hallucination of LMMs, consisting of realistic open-ended questions. It comprises 96 image-question pairs across 8 question categories and 12 object topics. The overall score is computed by GPT-4, which compares the model's response to the correct answer based on the given object information. If the overall score is less than 3, the response is considered to contain hallucinations.

\paragraph{Multimodal understanding benchmarks} We use MM-Vet \citep{yu2023mmvet} and MMBench \citep{liu2023mmbench} as benchmarks to measure the general multimodal performance of LMMs. MM-Vet is a benchmark consisting of 16 tasks and 218 instances designed to evaluate LMM's ability in complex multimodal tasks. The score is measured by GPT-4, which compares the LMM's response to the gold answer. MMBench comprises 4,377 multiple-choice questions aimed at assessing visual perception and visual reasoning. We utilize the development set of MMBench in this study.

\subsection{Baselines}
\label{subsec:baselines}
We use Openflamingo \citep{awadalla2023openflamingo}, MiniGPT-4 \citep{zhu2023minigpt4}, mPLUG-Owl \citep{ye2023mplugowl}, InstructBLIP \citep{dai2023instructblip}, Otter \citep{li2023otter}, LLaVA-SFT+ \citep{sun2023aligning}, and LLaVA-RLHF \citep{sun2023aligning} as baseline models. As multimodal hallucination corrector baselines, we employ LURE \citep{zhou2023analyzing} and Woodpecker \citep{yin2023woodpecker}. LURE utilizes MiniGPT-4 13B as its backbone model. Woodpecker uses gpt-3.5-turbo as its corrector, grounding DINO \citep{liu2023grounding} as its object detector and BLIP-2-FlanT5-XXL \citep{li2023blip2} for its VQA model.

\subsection{Main results}
\label{subsec:results}

\begin{table}[t]
\centering
\small
\begin{tabular}{lcc}
\toprule
\multirow{2}{*}{Model} & \multicolumn{2}{c}{MMHal-Bench} \\
& Score $\uparrow$ & Hal rate $\downarrow$  \\
\midrule
LURE & 1.9 & 0.58  \\
Woodpecker & 1.98 & 0.54  \\
{\Ours} 7B  & \textbf{2.6} & \textbf{0.49} \\
\midrule
LLaVA-RLHF 7B & 2.05 & 0.68  \\
{\Ours}\textsuperscript{--} 7B  & \textbf{2.19} & \textbf{0.59}  \\
\bottomrule
\end{tabular}
\caption{Performance comparison with recent methods focusing on reducing multimodal hallucination. {\Ours}\textsuperscript{--} 7B is a model fine-tuned with our multimodal feedback and revision data on LLaVA-SFT+ 7B, which is the backbone model of LLAVA-RLHF 7B.}

\label{tab:Table2}
\end{table}
\begin{table}[t]
\centering
\small
\begin{tabular}{lc|c}
\toprule
\multirow{2}{*}{Model} & \multicolumn{1}{c|}{MMBench} & \multicolumn{1}{c}{MM-Vet} \\
& Acc $\uparrow$ & Acc $\uparrow$ \\
\midrule
Openflamingo 9B & 6.6 & 24.8\\
MiniGPT-4 13B & 24.3 & 24.4 \\
InstructBLIP 14B & 36.0 & 25.6 \\
Otter 9B & 51.4	& 24.7  \\
LLaVA-SFT+ 7B & 52.7 & 30.4 \\
LLaVA-RLHF 7B & 52.7 & 29.8 \\
LLaVA-SFT+ 13B & 59.6 & 36.1 \\
LLaVA-RLHF 13B  & 59.6 & 36.4 \\
\midrule
LLaVA-1.5 7B & 59.9 & 31.2 \\
LLaVA-1.5 13B & 67.7 & 36.1 \\
{\Ours} 7B & 62.3 & 32.0 \\
{\Ours} 13B & \textbf{69.4} & \textbf{38.0} \\
\bottomrule
\end{tabular}
\caption{Results on multimodal understanding benchmarks. The detailed evaluation results for each benchmark by question type are in Table~\ref{tab:Table8} and Table~\ref{tab:Table9}.}
\label{tab:Table3}
\end{table}

\paragraph{{\Ours} achieves the best performance in multimodal hallucination benchmarks.} As shown in Table~\ref{tab:Table1}, {\Ours} consistently outperforms the base model, LLaVA-1.5 and other existing LMMs in the multimodal hallucination benchmark. It shows strong performance in benchmarks that measure scores using proprietary LLMs (MMHal-Bench, GAVIE) and a benchmark using conventional metrics like accuracy and F1 score (POPE). Notably, results from GAVIE demonstrate that {\Ours} not only provides accurate answers for a given image but also enhances its ability to follow instructions. Full results are in Appendix~\ref{subsec:hallucination_benchmarks_results}.

\paragraph{Natural language self-feedback is effective in revising responses.} Table~\ref{tab:Table2} shows {\Ours}'s effectiveness by comparing it with previous models designed to tackle multimodal hallucination. Compared to LURE and Woodpecker, both of which revise responses without feedback, {\Ours} reduces hallucination better. This suggests that providing specific feedback is crucial for correcting multimodal hallucination. In addition, unlike the two methods that require a separate model specialized for revision, {\Ours} efficiently gives better responses using just one model. Another notable observation is that Woodpecker's improvement in hallucination is less significant compared to {\Ours}, despite converting visual information into text and feeding it to a proprietary LLM corrector. From this, we find that for reducing multimodal hallucination, conveying visual features directly to the corrector model is critical. 

Compared to LLaVA-RLHF, which reduces LLM hallucination using RLHF, {\Ours} consistently performs better as well. For a fair comparison, we developed {\Ours}\textsuperscript{--} 7B by fine-tuning the base model of LLaVA-RLHF 7B, LLaVA-SFT+ 7B, on our multimodal feedback and revision data. The results indicate that providing feedback in the form of natural language feedback, which the model can directly interpret, is more effective than providing feedback as scalar values.

\paragraph{{\Ours} showcases high general multimodal understanding capabilities.} As the tendency of hallucination decreases, it is expected that the LMM can answer user questions about images more accurately. In this sense, we anticipate that {\Ours} would score high in benchmarks measuring general LMM's performance. To demonstrate this, we evaluate {\Ours} on benchmarks assessing complicated visual reasoning and perception capabilities of LLMs (Table~\ref{tab:Table3}). {\Ours} achieves superior performance compared to existing LMMs. Notably, as shown in Table~\ref{tab:Table8}, when measuring the math score related to a model's arithmetic capability, {\Ours} 13B impressively scored about twice as high as LLaVA-1.5 13B. Full results are in Appendix~\ref{subsec:understanding_benchmarks_results}.

\begin{table}[t]
\centering
\small
\begin{tabular}{lcc}
\toprule
\multirow{2}{*}{Model} & \multicolumn{2}{c}{MMHal-Bench} \\
& Score $\uparrow$ & Hal rate $\downarrow$  \\
\midrule
Only prediction  & 2.45 & 0.52  \\
No decision & 2.33	& 0.56  \\
{\Ours} 7B & \textbf{2.6} & \textbf{0.49} \\
\bottomrule
\end{tabular}
\caption{Module ablation results. The "Only prediction" is the result of performing only stage 1 for {\Ours} 7B. "No decision" is the outcome of completing stages 1 and 2.}
\label{tab:Table4}
\end{table}

\begin{table}[t]
\centering
\small
\begin{tabular}{lcc}
\toprule
\multirow{2}{*}{Model} & \multicolumn{2}{c}{MMHal-Bench} \\
& Score $\uparrow$ & Hal rate $\downarrow$  \\
\midrule
Iter 1  & 2.54 & 0.51  \\
Iter 2 & 2.58 & 0.5  \\
Iter 3 ({\Ours} 7B) & \textbf{2.6} & \textbf{0.49} \\
\bottomrule
\end{tabular}
\caption{Iteration ablation results.}
\label{tab:Table5}
\end{table}

\subsection{Ablation studies}
\label{subsec:ablation}

\paragraph{Module ablation} We test the influence of each stage in the critique-revise-decide process (Section~\ref{subsec:iterative}) in reducing multimodal hallucination. As shown in Table~\ref{tab:Table4}, when we only use the initial response as the final response and skip iterative self-revision, it scores lower than going through both processes. Surprisingly, even after just completing stage 1 and without self-revision, it still scores higher than the base model LLaVA-1.5 7B. This shows that merely fine-tuning with multimodal feedback and revision data can effectively reduce the hallucination rate. We observe a decrease in performance when the revised response is given as the final output without executing stage 3, compared to when a decision is made. This highlights the role of stage 3 in decreasing hallucination as it can prevent unnecessary revisions. This also suggests that while it is hard for the model to produce the right answer initially, distinguishing between right and wrong answers is relatively easier. 

\paragraph{Iteration ablation} We test how changing the maximum number of iterations affects {\Ours}'s performance. As shown in Table~\ref{tab:Table5}, as the maximum iteration count increased, the hallucination rate decreased. This indicates that answers are successfully refined through multiple revisions. However, there also exists a trade-off: as the iteration count goes up, the inference time also increases.

\begin{figure*}[t]
    \centering
    \includegraphics[width=1\linewidth]{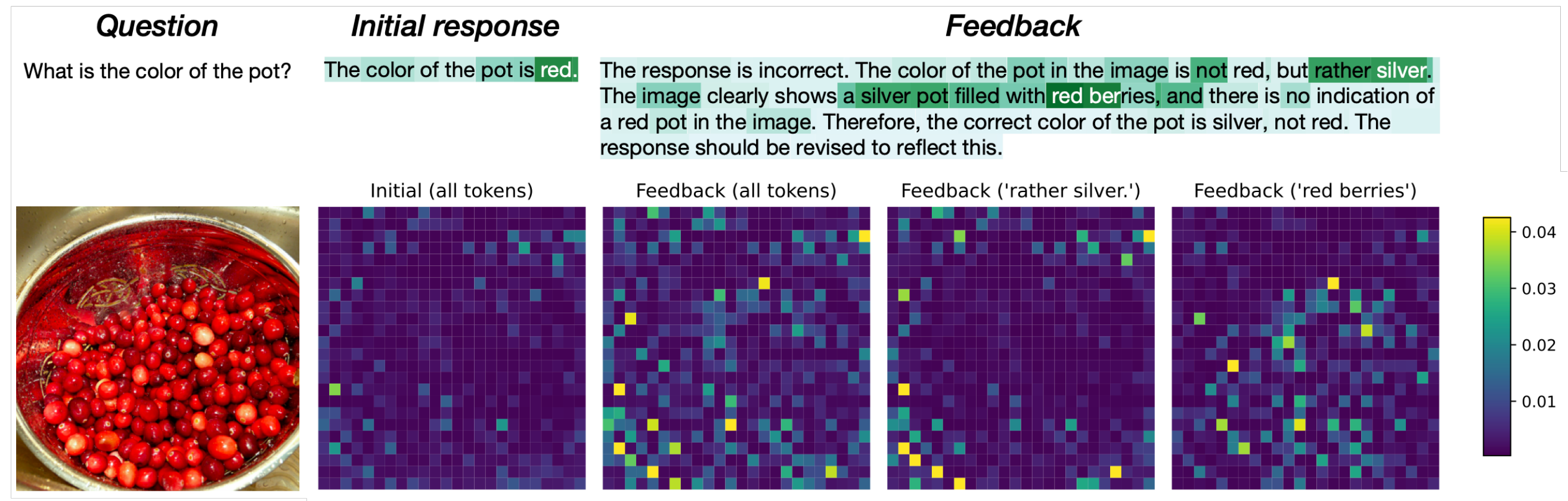}
    \caption{Coverage of image features attended during initial response and feedback generation on a single MMHal-Bench instance. The image attention heatmaps depict how the model's attention is distributed across image features, considering either all tokens or a subset of tokens in the output. In the text attention heatmaps above, the intensity of each token's background indicates the attention weight magnitude to image features, with darker highlights signifying higher weights. In the image attention heatmaps below, outliers at or above the 0.995th quantile are shown with the highest color intensity.}
    \label{fig:figure5}
\end{figure*}

\begin{figure}[ht]
    \centering
    \includegraphics[width=1\linewidth]{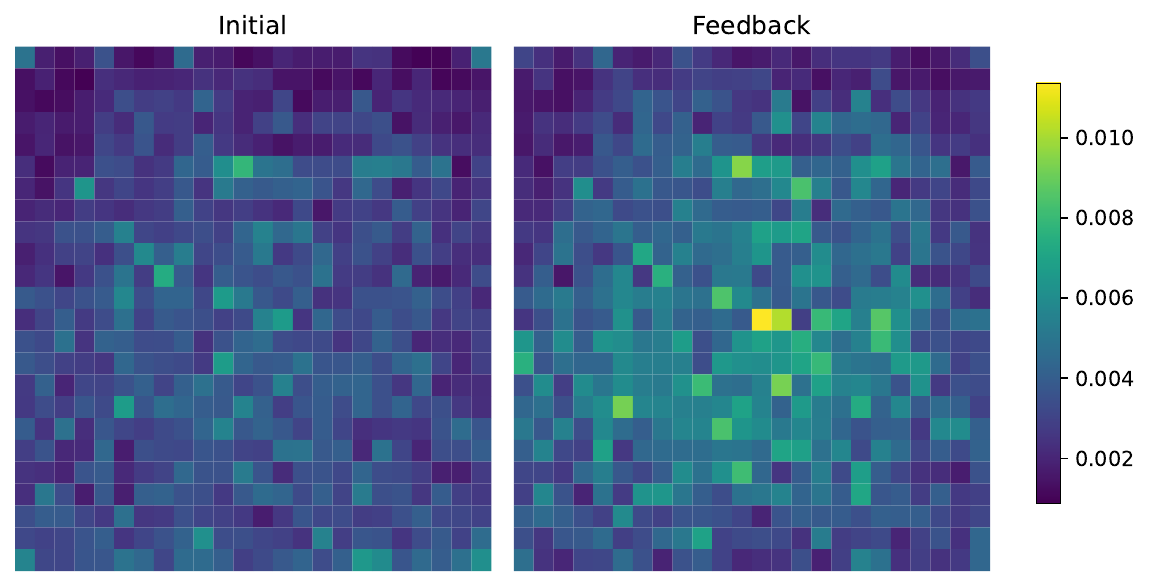}
    \caption{Average amount of attention to image features during the initial response (left) and feedback (right) generation. Attention weights are averaged across instances in MMHal-Bench where {\Ours}'s revision enhances the initial response.}
    \label{fig:figure4}
\end{figure}


\section{Qualitative analysis}
\label{sec:qualitative}
We qualitatively analyze how feedback from {\Ours} is effective in reducing multimodal hallucination. In this section, we examine {\Ours} 7B results on MMHal-Bench in which the model's revised answer is chosen as the final answer. We compare the visual information content between the model's initial response and feedback, focusing on amount (\ref{subsec:amount}) and coverage (\ref{subsec:coverage}). 
\subsection{Amount of visual information}
\label{subsec:amount}
Upon manual inspection of the instances, we observe that the initial response often correctly identifies object-level information but frequently misinterprets finer details such as object attributes or relationships between objects. On the contrary, we discover that the feedback text tends to describe the image contents more comprehensively.

To explore this phenomenon, we take inspiration from \citet{wang2023evaluation} and visualize how \textit{much} do output tokens attend to input image features\footnote{For every image, the vision encoder of {\Ours}, CLIP ViT-L/14 336px, processes it into 336px$\times$336px size and divides it into 14px$\times$14px patches, creating an image feature vector of size 576.} while generating initial response and feedback tokens to the same image. For each instance, we perform top-$k$ mean pooling to aggregate attention weights of initial response or feedback tokens on each image feature.\footnote{We experimented with min, max, mean, and top-$k$ mean pooling. We chose the top-$k$ mean configuration as it provided the clearest visualization for our analysis.} Specifically, we average top-3 attention weights across hidden layers, average top-3 weights across self-attention heads, and then average top-$l$ weights across output tokens, where $l$ is the length of the shorter output between initial response and feedback.

The results averaged across all instances are shown in Figure~\ref{fig:figure4}. Image features are more strongly attended by feedback than by initial response. Interestingly, even though attention to input would be more dispersed when generating feedback as its input includes an initial response in addition to the question, an increased concentration on wider areas of image features is visible. This suggests that visual information is largely reflected in the feedback text, supporting our manual inspection beforehand.

\subsection{Coverage of visual information}
\label{subsec:coverage}
We further empirically investigate how attention from individual tokens contributes to the \textit{coverage} of critical visual information. Building upon the visualization method described in Section~\ref{subsec:amount}, we compare attention weights to image features from \textit{all} tokens in the output with those from \textit{a subset of} tokens in the output. For the latter procedure, tokens that most intensely attend to image features during generation are deemed salient and are selected. 

We provide attention heatmaps of a representative instance in Figure~\ref{fig:figure5}. The task in this example is to identify the color of the pot in the image, and the initial response incorrectly answers ("red") and then the feedback corrects the answer ("silver"). 

Such a correction can be explained by the difference in the distribution of attention to image features during the generation of each token. Based on the heatmaps of all tokens attending to image features, when {\Ours} generates the initial response, it mostly focuses on features on the outer edges, corresponding to the rim of the pot; when generating feedback, it attends to the entire image including outer regions corresponding to the silver pot and inner regions with red berries in it. Heatmaps of specific tokens attending to image features show that in the process of improving the initial response, {\Ours} indeed focuses on the exact areas of the image corresponding to key color descriptors "silver" and "red" when generating these words. 

The findings suggest that during the feedback generation phase, the model develops an enhanced focus on an increased coverage of salient features, leading to a more comprehensive understanding of the image. This capability is beneficial for addressing the fundamental cause of multimodal hallucination of LLMs, which is that a lack of clear visual features leads LLMs to base their responses on pre-existing knowledge \citep{zhai2023halleswitch, li2023evaluating, liu2023mitigating, wang2023evaluation}. We propose that {\Ours}, with its ability to extract fine-grained visual information through feedback, can effectively reduce multimodal hallucination. 

\section{Conclusion}
\label{sec:conclusion}
In our work, we suggest a novel approach that utilizes feedback as visual signals to direct the model to refine responses that do not accurately reflect the image. Building on this approach, we present {\Ours}, a multimodal self-feedback guided revision model. {\Ours} has not only achieved state-of-the-art results on a multimodal hallucination benchmark but also demonstrated its effectiveness by improving performance compared to baseline models on multimodal understanding benchmarks. Through qualitative analysis, we demonstrate that the feedback produced by {\Ours} is well-grounded on the image, and providing the model with rich visual information helps reduce multimodal hallucination. We hope our model and data open new pathways for strategies to mitigate multimodal hallucination and uncover the fundamental cause of the issue.

\section*{Limitations}
\label{sec:limit}
In our study, we successfully demonstrated that {\Ours} can mitigate multimodal hallucination, as evidenced by our evaluations and analyses across various benchmarks. However, one notable drawback is the increased execution time. {\Ours} necessitates multiple calls to the model, making it less time-efficient than directly generating a response. On average, {\Ours} tends to be around 2 to 3 times slower than the base model, requiring 5.8 seconds to generate a response for a given image and instruction compared to 2.7 seconds by LLaVA-1.5. A strategy we use to reduce the overall execution time is to limit the number of iterations to 3. We think future work could explore improving the efficiency of the self-feedback-guided revision process. 

\section*{Acknowledgements}
We thank Seungone Kim, Seonghyun Ye, Doyoung Kim and Miyoung Ko for helpful discussion and valuable feedback on our work. This work was partly supported by Institute of Information \& communications Technology Planning \& Evaluation (IITP) grant funded by the Korea government (MSIT) (No.2022-0-00264, Comprehensive Video Understanding and Generation with Knowledge-based Deep Logic Neural Network, 80\%; No.2019-0-00075, Artificial Intelligence Graduate School Program (KAIST), 20\%).

\bibliography{acl_latex}

\begin{thebibliography}{50}
\expandafter\ifx\csname natexlab\endcsname\relax\def\natexlab#1{#1}\fi

\bibitem[{Akyürek et~al.(2023)Akyürek, Akyürek, Madaan, Kalyan, Clark, Wijaya, and Tandon}]{akyürek2023rl4f}
Afra~Feyza Akyürek, Ekin Akyürek, Aman Madaan, Ashwin Kalyan, Peter Clark, Derry Wijaya, and Niket Tandon. 2023.
\newblock \href {http://arxiv.org/abs/2305.08844} {Rl4f: Generating natural language feedback with reinforcement learning for repairing model outputs}.

\bibitem[{Alayrac et~al.(2022)Alayrac, Donahue, Luc, Miech, Barr, Hasson, Lenc, Mensch, Millican, Reynolds, Ring, Rutherford, Cabi, Han, Gong, Samangooei, Monteiro, Menick, Borgeaud, Brock, Nematzadeh, Sharifzadeh, Binkowski, Barreira, Vinyals, Zisserman, and Simonyan}]{alayrac2022flamingo}
Jean-Baptiste Alayrac, Jeff Donahue, Pauline Luc, Antoine Miech, Iain Barr, Yana Hasson, Karel Lenc, Arthur Mensch, Katie Millican, Malcolm Reynolds, Roman Ring, Eliza Rutherford, Serkan Cabi, Tengda Han, Zhitao Gong, Sina Samangooei, Marianne Monteiro, Jacob Menick, Sebastian Borgeaud, Andrew Brock, Aida Nematzadeh, Sahand Sharifzadeh, Mikolaj Binkowski, Ricardo Barreira, Oriol Vinyals, Andrew Zisserman, and Karen Simonyan. 2022.
\newblock \href {http://arxiv.org/abs/2204.14198} {Flamingo: a visual language model for few-shot learning}.

\bibitem[{Awadalla et~al.(2023)Awadalla, Gao, Gardner, Hessel, Hanafy, Zhu, Marathe, Bitton, Gadre, Sagawa, Jitsev, Kornblith, Koh, Ilharco, Wortsman, and Schmidt}]{awadalla2023openflamingo}
Anas Awadalla, Irena Gao, Josh Gardner, Jack Hessel, Yusuf Hanafy, Wanrong Zhu, Kalyani Marathe, Yonatan Bitton, Samir Gadre, Shiori Sagawa, Jenia Jitsev, Simon Kornblith, Pang~Wei Koh, Gabriel Ilharco, Mitchell Wortsman, and Ludwig Schmidt. 2023.
\newblock \href {http://arxiv.org/abs/2308.01390} {Openflamingo: An open-source framework for training large autoregressive vision-language models}.

\bibitem[{Biten et~al.(2022)Biten, Gómez, and Karatzas}]{9706727}
Ali~Furkan Biten, Lluís Gómez, and Dimosthenis Karatzas. 2022.
\newblock \href {https://doi.org/10.1109/WACV51458.2022.00253} {Let there be a clock on the beach: Reducing object hallucination in image captioning}.
\newblock In \emph{2022 IEEE/CVF Winter Conference on Applications of Computer Vision (WACV)}, pages 2473--2482.

\bibitem[{Chen et~al.(2023)Chen, Zhang, Zeng, Zhang, Zhu, and Zhao}]{chen2023shikra}
Keqin Chen, Zhao Zhang, Weili Zeng, Richong Zhang, Feng Zhu, and Rui Zhao. 2023.
\newblock \href {http://arxiv.org/abs/2306.15195} {Shikra: Unleashing multimodal llm's referential dialogue magic}.

\bibitem[{Dai et~al.(2023)Dai, Li, Li, Tiong, Zhao, Wang, Li, Fung, and Hoi}]{dai2023instructblip}
Wenliang Dai, Junnan Li, Dongxu Li, Anthony Meng~Huat Tiong, Junqi Zhao, Weisheng Wang, Boyang Li, Pascale Fung, and Steven Hoi. 2023.
\newblock \href {http://arxiv.org/abs/2305.06500} {Instructblip: Towards general-purpose vision-language models with instruction tuning}.

\bibitem[{Dubois et~al.(2023)Dubois, Li, Taori, Zhang, Gulrajani, Ba, Guestrin, Liang, and Hashimoto}]{dubois2023alpacafarm}
Yann Dubois, Xuechen Li, Rohan Taori, Tianyi Zhang, Ishaan Gulrajani, Jimmy Ba, Carlos Guestrin, Percy Liang, and Tatsunori~B. Hashimoto. 2023.
\newblock \href {http://arxiv.org/abs/2305.14387} {Alpacafarm: A simulation framework for methods that learn from human feedback}.

\bibitem[{Gou et~al.(2024)Gou, Shao, Gong, yelong shen, Yang, Duan, and Chen}]{gou2024critic}
Zhibin Gou, Zhihong Shao, Yeyun Gong, yelong shen, Yujiu Yang, Nan Duan, and Weizhu Chen. 2024.
\newblock \href {https://openreview.net/forum?id=Sx038qxjek} {{CRITIC}: Large language models can self-correct with tool-interactive critiquing}.
\newblock In \emph{The Twelfth International Conference on Learning Representations}.

\bibitem[{Gunjal et~al.(2023)Gunjal, Yin, and Bas}]{gunjal2023detecting}
Anisha Gunjal, Jihan Yin, and Erhan Bas. 2023.
\newblock \href {http://arxiv.org/abs/2308.06394} {Detecting and preventing hallucinations in large vision language models}.

\bibitem[{Ji et~al.(2023)Ji, Lee, Frieske, Yu, Su, Xu, Ishii, Bang, Madotto, and Fung}]{10.1145/3571730}
Ziwei Ji, Nayeon Lee, Rita Frieske, Tiezheng Yu, Dan Su, Yan Xu, Etsuko Ishii, Ye~Jin Bang, Andrea Madotto, and Pascale Fung. 2023.
\newblock \href {https://doi.org/10.1145/3571730} {Survey of hallucination in natural language generation}.
\newblock \emph{ACM Comput. Surv.}, 55(12).

\bibitem[{Kim et~al.(2023)Kim, Shin, Cho, Jang, Longpre, Lee, Yun, Shin, Kim, Thorne, and Seo}]{kim2023prometheus}
Seungone Kim, Jamin Shin, Yejin Cho, Joel Jang, Shayne Longpre, Hwaran Lee, Sangdoo Yun, Seongjin Shin, Sungdong Kim, James Thorne, and Minjoon Seo. 2023.
\newblock \href {http://arxiv.org/abs/2310.08491} {Prometheus: Inducing fine-grained evaluation capability in language models}.

\bibitem[{Lee et~al.(2023)Lee, Phatale, Mansoor, Lu, Mesnard, Bishop, Carbune, and Rastogi}]{lee2023rlaif}
Harrison Lee, Samrat Phatale, Hassan Mansoor, Kellie Lu, Thomas Mesnard, Colton Bishop, Victor Carbune, and Abhinav Rastogi. 2023.
\newblock \href {http://arxiv.org/abs/2309.00267} {Rlaif: Scaling reinforcement learning from human feedback with ai feedback}.

\bibitem[{Li et~al.(2023{\natexlab{a}})Li, Zhang, Chen, Wang, Yang, and Liu}]{li2023otter}
Bo~Li, Yuanhan Zhang, Liangyu Chen, Jinghao Wang, Jingkang Yang, and Ziwei Liu. 2023{\natexlab{a}}.
\newblock \href {http://arxiv.org/abs/2305.03726} {Otter: A multi-modal model with in-context instruction tuning}.

\bibitem[{Li et~al.(2023{\natexlab{b}})Li, Li, Savarese, and Hoi}]{li2023blip2}
Junnan Li, Dongxu Li, Silvio Savarese, and Steven Hoi. 2023{\natexlab{b}}.
\newblock \href {http://arxiv.org/abs/2301.12597} {Blip-2: Bootstrapping language-image pre-training with frozen image encoders and large language models}.

\bibitem[{Li et~al.(2023{\natexlab{c}})Li, Cheng, Zhao, Nie, and Wen}]{li2023halueval}
Junyi Li, Xiaoxue Cheng, Wayne~Xin Zhao, Jian-Yun Nie, and Ji-Rong Wen. 2023{\natexlab{c}}.
\newblock \href {http://arxiv.org/abs/2305.11747} {Halueval: A large-scale hallucination evaluation benchmark for large language models}.

\bibitem[{Li et~al.(2023{\natexlab{d}})Li, Du, Zhou, Wang, Zhao, and Wen}]{li2023evaluating}
Yifan Li, Yifan Du, Kun Zhou, Jinpeng Wang, Wayne~Xin Zhao, and Ji-Rong Wen. 2023{\natexlab{d}}.
\newblock \href {http://arxiv.org/abs/2305.10355} {Evaluating object hallucination in large vision-language models}.

\bibitem[{Lightman et~al.(2023)Lightman, Kosaraju, Burda, Edwards, Baker, Lee, Leike, Schulman, Sutskever, and Cobbe}]{lightman2023lets}
Hunter Lightman, Vineet Kosaraju, Yura Burda, Harri Edwards, Bowen Baker, Teddy Lee, Jan Leike, John Schulman, Ilya Sutskever, and Karl Cobbe. 2023.
\newblock \href {http://arxiv.org/abs/2305.20050} {Let's verify step by step}.

\bibitem[{Liu et~al.(2023{\natexlab{a}})Liu, Lin, Li, Wang, Yacoob, and Wang}]{liu2023mitigating}
Fuxiao Liu, Kevin Lin, Linjie Li, Jianfeng Wang, Yaser Yacoob, and Lijuan Wang. 2023{\natexlab{a}}.
\newblock \href {http://arxiv.org/abs/2306.14565} {Mitigating hallucination in large multi-modal models via robust instruction tuning}.

\bibitem[{Liu et~al.(2023{\natexlab{b}})Liu, Li, Li, and Lee}]{liu2023improved}
Haotian Liu, Chunyuan Li, Yuheng Li, and Yong~Jae Lee. 2023{\natexlab{b}}.
\newblock \href {http://arxiv.org/abs/2310.03744} {Improved baselines with visual instruction tuning}.

\bibitem[{Liu et~al.(2023{\natexlab{c}})Liu, Li, Wu, and Lee}]{liu2023visual}
Haotian Liu, Chunyuan Li, Qingyang Wu, and Yong~Jae Lee. 2023{\natexlab{c}}.
\newblock Visual instruction tuning.
\newblock \emph{arXiv preprint arXiv:2304.08485}.

\bibitem[{Liu et~al.(2023{\natexlab{d}})Liu, Zeng, Ren, Li, Zhang, Yang, Li, Yang, Su, Zhu, and Zhang}]{liu2023grounding}
Shilong Liu, Zhaoyang Zeng, Tianhe Ren, Feng Li, Hao Zhang, Jie Yang, Chunyuan Li, Jianwei Yang, Hang Su, Jun Zhu, and Lei Zhang. 2023{\natexlab{d}}.
\newblock \href {http://arxiv.org/abs/2303.05499} {Grounding dino: Marrying dino with grounded pre-training for open-set object detection}.

\bibitem[{Liu et~al.(2023{\natexlab{e}})Liu, Duan, Zhang, Li, Zhang, Zhao, Yuan, Wang, He, Liu, Chen, and Lin}]{liu2023mmbench}
Yuan Liu, Haodong Duan, Yuanhan Zhang, Bo~Li, Songyang Zhang, Wangbo Zhao, Yike Yuan, Jiaqi Wang, Conghui He, Ziwei Liu, Kai Chen, and Dahua Lin. 2023{\natexlab{e}}.
\newblock \href {http://arxiv.org/abs/2307.06281} {Mmbench: Is your multi-modal model an all-around player?}

\bibitem[{Ma et~al.(2023)Ma, Liang, Wang, Huang, Bastani, Jayaraman, Zhu, Fan, and Anandkumar}]{ma2023eureka}
Yecheng~Jason Ma, William Liang, Guanzhi Wang, De-An Huang, Osbert Bastani, Dinesh Jayaraman, Yuke Zhu, Linxi Fan, and Anima Anandkumar. 2023.
\newblock \href {http://arxiv.org/abs/2310.12931} {Eureka: Human-level reward design via coding large language models}.

\bibitem[{Maaz et~al.(2023)Maaz, Rasheed, Khan, and Khan}]{maaz2023videochatgpt}
Muhammad Maaz, Hanoona Rasheed, Salman Khan, and Fahad~Shahbaz Khan. 2023.
\newblock \href {http://arxiv.org/abs/2306.05424} {Video-chatgpt: Towards detailed video understanding via large vision and language models}.

\bibitem[{Madaan et~al.(2023)Madaan, Tandon, Gupta, Hallinan, Gao, Wiegreffe, Alon, Dziri, Prabhumoye, Yang, Gupta, Majumder, Hermann, Welleck, Yazdanbakhsh, and Clark}]{madaan2023selfrefine}
Aman Madaan, Niket Tandon, Prakhar Gupta, Skyler Hallinan, Luyu Gao, Sarah Wiegreffe, Uri Alon, Nouha Dziri, Shrimai Prabhumoye, Yiming Yang, Shashank Gupta, Bodhisattwa~Prasad Majumder, Katherine Hermann, Sean Welleck, Amir Yazdanbakhsh, and Peter Clark. 2023.
\newblock \href {http://arxiv.org/abs/2303.17651} {Self-refine: Iterative refinement with self-feedback}.

\bibitem[{OpenAI(2022)}]{chatgpt}
OpenAI. 2022.
\newblock \href {https://openai.com/blog/chatgpt/} {Chatgpt: Optimizing language models for dialogue}.

\bibitem[{Ouyang et~al.(2022)Ouyang, Wu, Jiang, Almeida, Wainwright, Mishkin, Zhang, Agarwal, Slama, Ray, Schulman, Hilton, Kelton, Miller, Simens, Askell, Welinder, Christiano, Leike, and Lowe}]{ouyang2022training}
Long Ouyang, Jeff Wu, Xu~Jiang, Diogo Almeida, Carroll~L. Wainwright, Pamela Mishkin, Chong Zhang, Sandhini Agarwal, Katarina Slama, Alex Ray, John Schulman, Jacob Hilton, Fraser Kelton, Luke Miller, Maddie Simens, Amanda Askell, Peter Welinder, Paul Christiano, Jan Leike, and Ryan Lowe. 2022.
\newblock \href {http://arxiv.org/abs/2203.02155} {Training language models to follow instructions with human feedback}.

\bibitem[{Pan et~al.(2023)Pan, Saxon, Xu, Nathani, Wang, and Wang}]{pan2023automatically}
Liangming Pan, Michael Saxon, Wenda Xu, Deepak Nathani, Xinyi Wang, and William~Yang Wang. 2023.
\newblock \href {http://arxiv.org/abs/2308.03188} {Automatically correcting large language models: Surveying the landscape of diverse self-correction strategies}.

\bibitem[{Peng et~al.(2023)Peng, Wang, Dong, Hao, Huang, Ma, and Wei}]{peng2023kosmos2}
Zhiliang Peng, Wenhui Wang, Li~Dong, Yaru Hao, Shaohan Huang, Shuming Ma, and Furu Wei. 2023.
\newblock \href {http://arxiv.org/abs/2306.14824} {Kosmos-2: Grounding multimodal large language models to the world}.

\bibitem[{Rohrbach et~al.(2018)Rohrbach, Hendricks, Burns, Darrell, and Saenko}]{rohrbach-etal-2018-object}
Anna Rohrbach, Lisa~Anne Hendricks, Kaylee Burns, Trevor Darrell, and Kate Saenko. 2018.
\newblock \href {https://doi.org/10.18653/v1/D18-1437} {Object hallucination in image captioning}.
\newblock In \emph{Proceedings of the 2018 Conference on Empirical Methods in Natural Language Processing}, pages 4035--4045, Brussels, Belgium. Association for Computational Linguistics.

\bibitem[{Scheurer et~al.(2022)Scheurer, Campos, Chan, Chen, Cho, and Perez}]{scheurer2022training}
Jérémy Scheurer, Jon~Ander Campos, Jun~Shern Chan, Angelica Chen, Kyunghyun Cho, and Ethan Perez. 2022.
\newblock \href {http://arxiv.org/abs/2204.14146} {Training language models with language feedback}.

\bibitem[{Shinn et~al.(2023)Shinn, Cassano, Berman, Gopinath, Narasimhan, and Yao}]{shinn2023reflexion}
Noah Shinn, Federico Cassano, Edward Berman, Ashwin Gopinath, Karthik Narasimhan, and Shunyu Yao. 2023.
\newblock \href {http://arxiv.org/abs/2303.11366} {Reflexion: Language agents with verbal reinforcement learning}.

\bibitem[{Su et~al.(2023)Su, Lan, Li, Xu, Wang, and Cai}]{su2023pandagpt}
Yixuan Su, Tian Lan, Huayang Li, Jialu Xu, Yan Wang, and Deng Cai. 2023.
\newblock \href {http://arxiv.org/abs/2305.16355} {Pandagpt: One model to instruction-follow them all}.

\bibitem[{Sun et~al.(2023)Sun, Shen, Cao, Liu, Li, Shen, Gan, Gui, Wang, Yang, Keutzer, and Darrell}]{sun2023aligning}
Zhiqing Sun, Sheng Shen, Shengcao Cao, Haotian Liu, Chunyuan Li, Yikang Shen, Chuang Gan, Liang-Yan Gui, Yu-Xiong Wang, Yiming Yang, Kurt Keutzer, and Trevor Darrell. 2023.
\newblock \href {http://arxiv.org/abs/2309.14525} {Aligning large multimodal models with factually augmented rlhf}.

\bibitem[{Wang et~al.(2023{\natexlab{a}})Wang, Wu, Han, Peng, Zhong, Zhang, Dong, Li, Li, Wang, and He}]{wang2023vigc}
Bin Wang, Fan Wu, Xiao Han, Jiahui Peng, Huaping Zhong, Pan Zhang, Xiaoyi Dong, Weijia Li, Wei Li, Jiaqi Wang, and Conghui He. 2023{\natexlab{a}}.
\newblock \href {http://arxiv.org/abs/2308.12714} {Vigc: Visual instruction generation and correction}.

\bibitem[{Wang et~al.(2023{\natexlab{b}})Wang, Zhou, Xu, Shi, Zhao, Xu, Ye, Yan, Zhang, Zhu, Sang, and Tang}]{wang2023evaluation}
Junyang Wang, Yiyang Zhou, Guohai Xu, Pengcheng Shi, Chenlin Zhao, Haiyang Xu, Qinghao Ye, Ming Yan, Ji~Zhang, Jihua Zhu, Jitao Sang, and Haoyu Tang. 2023{\natexlab{b}}.
\newblock \href {http://arxiv.org/abs/2308.15126} {Evaluation and analysis of hallucination in large vision-language models}.

\bibitem[{Wang et~al.(2023{\natexlab{c}})Wang, Li, Chen, Cai, Zhu, Lin, Cao, Liu, Liu, and Sui}]{wang2023large}
Peiyi Wang, Lei Li, Liang Chen, Zefan Cai, Dawei Zhu, Binghuai Lin, Yunbo Cao, Qi~Liu, Tianyu Liu, and Zhifang Sui. 2023{\natexlab{c}}.
\newblock \href {http://arxiv.org/abs/2305.17926} {Large language models are not fair evaluators}.

\bibitem[{Wang et~al.(2023{\natexlab{d}})Wang, Yu, Tan, O'Brien, Pasunuru, Dwivedi-Yu, Golovneva, Zettlemoyer, Fazel-Zarandi, and Celikyilmaz}]{wang2023shepherd}
Tianlu Wang, Ping Yu, Xiaoqing~Ellen Tan, Sean O'Brien, Ramakanth Pasunuru, Jane Dwivedi-Yu, Olga Golovneva, Luke Zettlemoyer, Maryam Fazel-Zarandi, and Asli Celikyilmaz. 2023{\natexlab{d}}.
\newblock \href {http://arxiv.org/abs/2308.04592} {Shepherd: A critic for language model generation}.

\bibitem[{Welleck et~al.(2022)Welleck, Lu, West, Brahman, Shen, Khashabi, and Choi}]{welleck2022generating}
Sean Welleck, Ximing Lu, Peter West, Faeze Brahman, Tianxiao Shen, Daniel Khashabi, and Yejin Choi. 2022.
\newblock \href {http://arxiv.org/abs/2211.00053} {Generating sequences by learning to self-correct}.

\bibitem[{Wu et~al.(2023)Wu, Hu, Shi, Dziri, Suhr, Ammanabrolu, Smith, Ostendorf, and Hajishirzi}]{wu2023finegrained}
Zeqiu Wu, Yushi Hu, Weijia Shi, Nouha Dziri, Alane Suhr, Prithviraj Ammanabrolu, Noah~A. Smith, Mari Ostendorf, and Hannaneh Hajishirzi. 2023.
\newblock \href {http://arxiv.org/abs/2306.01693} {Fine-grained human feedback gives better rewards for language model training}.

\bibitem[{Ye et~al.(2023{\natexlab{a}})Ye, Xu, Xu, Ye, Yan, Zhou, Wang, Hu, Shi, Shi, Li, Xu, Chen, Tian, Qi, Zhang, and Huang}]{ye2023mplugowl}
Qinghao Ye, Haiyang Xu, Guohai Xu, Jiabo Ye, Ming Yan, Yiyang Zhou, Junyang Wang, Anwen Hu, Pengcheng Shi, Yaya Shi, Chenliang Li, Yuanhong Xu, Hehong Chen, Junfeng Tian, Qian Qi, Ji~Zhang, and Fei Huang. 2023{\natexlab{a}}.
\newblock \href {http://arxiv.org/abs/2304.14178} {mplug-owl: Modularization empowers large language models with multimodality}.

\bibitem[{Ye et~al.(2023{\natexlab{b}})Ye, Jo, Kim, Kim, Hwang, and Seo}]{selfee2023}
Seonghyeon Ye, Yongrae Jo, Doyoung Kim, Sungdong Kim, Hyeonbin Hwang, and Minjoon Seo. 2023{\natexlab{b}}.
\newblock \href {https://kaistai.github.io/SelFee/} {Selfee: Iterative self-revising llm empowered by self-feedback generation}.
\newblock Blog post.

\bibitem[{Yin et~al.(2023)Yin, Fu, Zhao, Xu, Wang, Sui, Shen, Li, Sun, and Chen}]{yin2023woodpecker}
Shukang Yin, Chaoyou Fu, Sirui Zhao, Tong Xu, Hao Wang, Dianbo Sui, Yunhang Shen, Ke~Li, Xing Sun, and Enhong Chen. 2023.
\newblock \href {http://arxiv.org/abs/2310.16045} {Woodpecker: Hallucination correction for multimodal large language models}.

\bibitem[{Yu et~al.(2023)Yu, Yang, Li, Wang, Lin, Liu, Wang, and Wang}]{yu2023mmvet}
Weihao Yu, Zhengyuan Yang, Linjie Li, Jianfeng Wang, Kevin Lin, Zicheng Liu, Xinchao Wang, and Lijuan Wang. 2023.
\newblock \href {http://arxiv.org/abs/2308.02490} {Mm-vet: Evaluating large multimodal models for integrated capabilities}.

\bibitem[{Zhai et~al.(2023)Zhai, Yang, Zhao, Xu, Shen, Zhao, Keutzer, Li, Yan, and Fan}]{zhai2023halleswitch}
Bohan Zhai, Shijia Yang, Xiangchen Zhao, Chenfeng Xu, Sheng Shen, Dongdi Zhao, Kurt Keutzer, Manling Li, Tan Yan, and Xiangjun Fan. 2023.
\newblock \href {http://arxiv.org/abs/2310.01779} {Halle-switch: Rethinking and controlling object existence hallucinations in large vision language models for detailed caption}.

\bibitem[{Zhang et~al.(2023{\natexlab{a}})Zhang, Press, Merrill, Liu, and Smith}]{zhang2023language}
Muru Zhang, Ofir Press, William Merrill, Alisa Liu, and Noah~A. Smith. 2023{\natexlab{a}}.
\newblock \href {http://arxiv.org/abs/2305.13534} {How language model hallucinations can snowball}.

\bibitem[{Zhang et~al.(2023{\natexlab{b}})Zhang, Han, Liu, Gao, Zhou, Hu, Yan, Lu, Li, and Qiao}]{zhang2023llamaadapter}
Renrui Zhang, Jiaming Han, Chris Liu, Peng Gao, Aojun Zhou, Xiangfei Hu, Shilin Yan, Pan Lu, Hongsheng Li, and Yu~Qiao. 2023{\natexlab{b}}.
\newblock \href {http://arxiv.org/abs/2303.16199} {Llama-adapter: Efficient fine-tuning of language models with zero-init attention}.

\bibitem[{Zhang et~al.(2023{\natexlab{c}})Zhang, Li, Cui, Cai, Liu, Fu, Huang, Zhao, Zhang, Chen, Wang, Luu, Bi, Shi, and Shi}]{zhang2023sirens}
Yue Zhang, Yafu Li, Leyang Cui, Deng Cai, Lemao Liu, Tingchen Fu, Xinting Huang, Enbo Zhao, Yu~Zhang, Yulong Chen, Longyue Wang, Anh~Tuan Luu, Wei Bi, Freda Shi, and Shuming Shi. 2023{\natexlab{c}}.
\newblock \href {http://arxiv.org/abs/2309.01219} {Siren's song in the ai ocean: A survey on hallucination in large language models}.

\bibitem[{Zhou et~al.(2023)Zhou, Cui, Yoon, Zhang, Deng, Finn, Bansal, and Yao}]{zhou2023analyzing}
Yiyang Zhou, Chenhang Cui, Jaehong Yoon, Linjun Zhang, Zhun Deng, Chelsea Finn, Mohit Bansal, and Huaxiu Yao. 2023.
\newblock \href {http://arxiv.org/abs/2310.00754} {Analyzing and mitigating object hallucination in large vision-language models}.

\bibitem[{Zhu et~al.(2023)Zhu, Chen, Shen, Li, and Elhoseiny}]{zhu2023minigpt4}
Deyao Zhu, Jun Chen, Xiaoqian Shen, Xiang Li, and Mohamed Elhoseiny. 2023.
\newblock \href {http://arxiv.org/abs/2304.10592} {Minigpt-4: Enhancing vision-language understanding with advanced large language models}.

\end{thebibliography}

\appendix
\section{Full results on benchmarks}
\label{sec:full_results}
In this section, we describe the detailed results from the benchmarks used in our work. The benchmarks are designed to evaluate the performance of LMMs from multiple perspectives, encompassing various sub-tasks and types of questions. 

\subsection{Multimodal hallucination benchmarks}
\label{subsec:hallucination_benchmarks_results}
For MMHal-Bench, the questions are categorized into 8 types: Attribute, Adversarial, Comparison, Counting, Relation, Environment, Holistic, and Other (Table~\ref{tab:Table6}). POPE evaluates three types of questions: random, popular, and adversarial (Table~\ref{tab:Table7}). 

\subsection{Multimodal understanding benchmarks}
\label{subsec:understanding_benchmarks_results}
MM-Vet is composed of sub-tasks designed to measure 6 LMM capabilities: Recognition, OCR (Optical Character Recognition), Knowledge, Language generation, Spatial awareness, and Math (Table~\ref{tab:Table8}). MMBench is structured to evaluate across L-1, L-2, and L-3 dimensions. We followed previous works and conducted evaluations for the L-2 dimension. The L-2 dimension tasks include Coarse Perception (CP), Fine-grained Single-instance Perception (FP-S), Fine-grained Cross-instance Perception (FP-C), Attribute Reasoning (AR), Relation Reasoning (RR), and Logic Reasoning (LR) (Table~\ref{tab:Table9}).

\section{Prompts}

\subsection{Prompts for inference at each stage} 
\label{subsec:inference_prompts}
For all prompts, we did not explicitly provide an image feature prompt. Instead, the image features are concatenated with the question during the tokenization process before being input to the model. Additionally, the prompt for the decision process is based on the work of \citep{liu2023improved}.

\subsection{Prompt for generating multimodal feedback}
\label{subsec:data_prompts}
We introduce the prompt used in generating our multimodal feedback dataset. For an LLM that cannot see images, we included the image contents in the form of text within the prompt, allowing it to provide feedback as if it had seen the image and initial response. We utilized object information and a gold caption as the image contents. In instances where no objects are present in the dataset, we didn't use a separate object detector to prevent the model's errors from propagating into the feedback. Instead, only the gold caption is provided in such cases. Additionally, to avoid erroneously generating feedback that suggests the presence of hallucination merely due to the use of different expressions, even when the initial response aligns sufficiently with the image information but uses different terms from the gold answer, we crafted the prompt to treat synonyms or paraphrases as correct answers. Drawing inspiration from previous research \citep{kim2023prometheus}, we structured the prompt to ensure that it encapsulates these aspects well.

\section{Computation}
\label{sec:computation}
For this research, we used an NVIDIA A100-SXM4-80GB GPU and an AMD EPYC 7513 32-Core Processor running at 2.0778 GHz. Training {\Ours} 7B required 8 GPUs and took a total of 15 hours, while training {\Ours} 13B took 30 hours. While the time taken to evaluate each dataset varies, {\Ours} takes about 2 to 3 times longer to complete the entire process compared to existing baselines that only generate responses. 

\section{Hyperparameters}
\label{sec:hyperparameters}
We used a batch size of 128, a learning rate of 2e-5, and trained for 1 epoch. The maximum length is set to 2048, with no weight decay. We employed a cosine scheduler for learning rate adjustments, with a warmup ratio of 0.03. Additionally, we incorporated gradient checkpointing and used DeepSpeed zero stage 3. The maximum number of iterations for self-revision is 3. When generating responses, we utilized greedy decoding following LLaVA-1.5.
\begin{table*}[t]
\tiny
\centering
\begin{tabular}{lcccccccc|cc}
\toprule
Model  & Attribute $\uparrow$ &	Adversarial $\uparrow$ & 	Comparison $\uparrow$ &	Counting $\uparrow$ &	Relation $\uparrow$ &	Environment $\uparrow$	& Holistic	$\uparrow$ & Other $\uparrow$ & Score $\uparrow$ & Hal rate $\downarrow$\\
\midrule
Kosmos-2	&	2	&0.25&	1.42&	1.67&	1.67&	2.67&	2.5&	1.33 & 1.69& 	0.68\\
IDEFIC 9B	&	1.58&	0.75&	2.75&	1.83&	1.83&	2.5&	2.17&	1.67 &1.89&	0.64\\
IDEFIC 80B	&	2.33&	1.25&	2&	2.5&	1.5&	3.33&	2.33&	1.17 &2.05&	0.61\\
InstructBLIP 7B	&	3.42&	2.08&	1.33&	1.92&	2.17&	3.67&	1.17&	1.08 &2.1	&0.58\\
InstructBLIP 13B&	2.75&	1.75&	1.25&	2.08&	2.5&	\textbf{4.08}&	1.5&	1.17 &	2.14	&0.58\\
LLaVA-SFT+ 7B&	2.75&	2.08&	1.42&	1.83&	2.17&	2.17&	1.17&	0.5 &	1.76&	0.67\\
LLaVA-RLHF 7B&	2.92&	1.83&	2.42&	1.92&	2.25&	2.25&	1.75&	1.08 &	2.05&	0.68\\
LLaVA-SFT+ 13B&	3.08&	1.75&	2&	\textbf{3.25}&	2.25&	3.83&	1.5&	1.75 & 2.43&	0.55\\
LLaVA-RLHF 13B&	3.33&	\textbf{2.67}&	1.75&	2.25&	2.33&	3.25&	2.25&	\textbf{2.42} &	2.53&	0.57\\
\midrule
LLaVA-1.5 7B&	3.17&	1.25&	3.17&	2.5&	2.33&	3.17&	1.5&	2.25 &	2.42&	0.55\\
LLaVA-1.5 13B&	\textbf{3.5}&	2&	2.67&	2.33&	1.67&	3.33&	2.58&	2.25 &	2.54&	0.52\\
{\Ours} 7B&	3.42&	2.42&	3.08&	1.75&	\textbf{2.75}&	3.75&	1.33&	2.33 &	2.6&	0.49\\
{\Ours} 13B&	3&	1.75&	\textbf{3.42}&	1.67&	2.33&	3.75&	\textbf{2.75}&	\textbf{2.42} &	\textbf{2.64}&	\textbf{0.48}\\
\bottomrule
\end{tabular}
\caption{Results on MMHal-Bench}
\label{tab:Table6}
\end{table*}

\begin{table*}[t]
\scriptsize
\centering
\begin{tabular}{lccc|ccc|ccc|cc}
\toprule
\multirow{2}{*}{Model}  & \multicolumn{3}{c|}{Random} &	\multicolumn{3}{c|}{Popular} & \multicolumn{3}{c|}{Adversarial} & \multicolumn{2}{c}{Overall}\\
& Acc $\uparrow$ & F1 $\uparrow$ & Yes (\%) & Acc $\uparrow$ & F1 $\uparrow$ & Yes (\%) & Acc $\uparrow$ & F1 $\uparrow$ & Yes (\%) & Acc $\uparrow$ & F1 $\uparrow$ \\
\midrule
Shikra & 86.9 & 86.2 & 43.3 & 84 & 83.2 & 45.2 & 83.1 & 82.5 & 46.5 & 84.7 & 84.0 \\
InstructBLIP & 88.6 & 89.3 & 56.6	& 79.7	& 80.2	& 52.5&	65.2	&70.4	&67.8	&77.8	&80.0\\
MiniGPT-4&	79.7	&80.2&	52.5	&69.7&	73	&62.2&	65.2&	70.4&	67.8&	71.5	&74.5\\
mPLUG-Owl&	54	&68.4	&95.6&	50.9	&66.9&	98.6&	50.7&	66.8	&98.7&	51.9	&67.2\\
LLaVA-SFT+ 7B&	86.1	&85.5	&44.5	&82.9&	82.4	&47.2	&80.2&	80.1&	49.6	&83.1&	82.7\\
LLaVA-RLHF 7B&	84.8&	83.3&	39.6&	83.3	&81.8&	41.8	&80.7	&79.5	&44	&82.9&	81.5 \\
LLaVA-SFT+ 13B&	86&	84.8&	40.5&	84&	82.6&	41.6&	82.3	&81.1&	43.5&	84.1&	82.8\\
LLaVA-RLHF 13B&	85.2	&83.5	&38.4&	83.9&	81.8	&38	&82.3&	80.5	&40.5&	83.8&	81.9 \\
\midrule
LLaVA-1.5 7B	&88.2	&87.3&	41.9&	87.3	&86.2	&41.8&	85.2	&84.2	&44&	86.9	&85.9 \\
LLaVA-1.5 13B&	88&	87.1	&41.7&	87.4	&86.2	&41.3	&85.5&	84.5	&43.3	&87.0	&85.9 \\
{\Ours} 7B&	89.9	&89.4	&43.9	&\textbf{88.5}&	\textbf{87.9}	&45.1	&86.2&	85.7&	46.6	&88.2&	\textbf{87.7} \\
{\Ours} 13B& \textbf{90.2}& \textbf{89.7}& 44.3& 88.1& 87.4& 44.5& \textbf{86.6}& \textbf{86.1}& 46.7& \textbf{88.3}& \textbf{87.7}\\										
\bottomrule
\end{tabular}
\caption{Results on Pope}
\label{tab:Table7}
\end{table*}

\begin{table*}[t]
\centering
\begin{tabular}{lcccccc|c}
\toprule
Model & rec $\uparrow$ & ocr $\uparrow$& know $\uparrow$& gen $\uparrow$& spat $\uparrow$& math $\uparrow$& total $\uparrow$\\
\midrule
Transformers Agent (GPT-4)	& 18.2 & 3.9 & 2.2&	3.2	&12.4	&4&	13.4 \\
MiniGPT-4-8B&	27.4&	15&	12.8&	13.9	&20.3&	7.7&	22.1 \\
BLIP-2-12B	&27.5&	11.1	&11.8&	7&	16.2&	5.8	&22.4 \\
MiniGPT-4-14B&	29.9&	16.1&	20.4&	22.1&	22.2&	3.8&	24.4 \\
Otter-9B	&27.3&	17.8	&14.2&	13.8&	24.4&	3.8&	24.7 \\
OpenFlamingo-9B&	28.7&	16.7	&16.4	&13.1	&21	&7.7	&24.8 \\
InstructBLIP-14B&	30.8	&16	&9.8	&9	&21.1	&10.5	&25.6 \\
InstructBLIP-8B	&32.4	&14.6	&16.5	&18.2	&18.6	&7.7	&26.2 \\
LLaMA-Adapter v2-7B	3&8.5	&20.3	&\textbf{31.4}	&\textbf{33.4}	&22.9	&3.8	&31.4 \\
\midrule
LLaVA-1.5 7B	&37	&21	&17.6&	20.4&	24.9	&7.7&	31.2 \\
LLaVA-1.5 13B	&40.6	&28	&23.5	&24.4	&\textbf{34.7}	&7.7	&36.1 \\
{\Ours} 7B	&36.7	&23.5	&18.2	&22	&27.6	&3.8	&32 \\
{\Ours} 13B	&\textbf{42.9}	&\textbf{30.4}	&24.5	&29.2	&32.7	&\textbf{15}	&\textbf{38} \\
\bottomrule
\end{tabular}
\caption{Results on MM-Vet}
\label{tab:Table8}
\end{table*}

\begin{table*}[t]
\centering
\begin{tabular}{lcccccc|c}
\toprule
Model & LR $\uparrow$ &	AR $\uparrow$	& RR $\uparrow$	& FP-S $\uparrow$& 	FP-C $\uparrow$&	CP $\uparrow$& 	Overall $\uparrow$\\
\midrule
OpenFlamingo & 6.7 & 8 & 0 & 6.7 & 2.8	& 2	& 4.6 \\
OpenFlamingo v2	& 4.2	& 15.4	& 0.9	& 8.1	& 1.4	& 5	& 6.6 \\
MMGPT	& 2.5	& 26.4	& 13	& 14.1	& 3.4	& 20.8	& 15.3 \\
VisualGLM	& 10.8	& 44.3	& 35.7	& 43.8	& 23.4	& 47.3	& 38.1 \\
LLaMA-Adapter	& 11.7	& 35.3	& 29.6	& 47.5	& 38.6	& 56.4	& 41.2 \\
µ-G2PT	& 13.3	& 38.8	& 40.9	& 46.5	& 38.6	& 58.1	& 43.2 \\
mPLUG-Owl	& 16.7	& 53.2	& 47.8	& 50.2	& 40.7	& 64.1	& 49.4 \\
Otter	& 32.5	& 56.7	& 53.9	& 46.8	& 38.6	& 65.4	& 51.4 \\
Shikra	& 25.8	& 56.7	& 58.3	& 57.2	& 57.9	& 75.8	& 58.8 \\
Kosmos-2	& \textbf{46.7}	& 55.7	& 43.5	& 64.3	& 49	& 72.5	& 59.2 \\
PandaGPT	& 10	& 38.8	& 23.5	& 27.9	& 35.2	& 48.3	& 33.5 \\
MiniGPT-4	& 20.8	& 50.7	& 30.4	& 49.5	& 26.2	& 50.7	& 42.3 \\
InstructBLIP	& 19.1	& 54.2	& 34.8	& 47.8	& 24.8	& 56.4	& 44 \\
\midrule
LLaVA-1.5 7B	& 30.8	& \textbf{73.1}	& 53.9	& 67	& 57.2	& 77.2	& 59.9 \\
LLaVA-1.5 13B	& 41.7	& 69.7	& 63.5	& 70	& 59.3	& 80.2	& 67.7 \\
{\Ours} 7B	& 30.8	& 65.2	& 59.1	& 67.7	& 54.5	& 72.8	& 62.3 \\
{\Ours} 13B	& 38.3	& 70.6	&\textbf{67}	& \textbf{72.4}	& \textbf{62.8}	& \textbf{82.2}	& \textbf{69.4} \\
\bottomrule
\end{tabular}
\caption{Results on MMBench}
\label{tab:Table9}
\end{table*}
\begin{figure*}[h]
\includegraphics[width=1.0\linewidth]{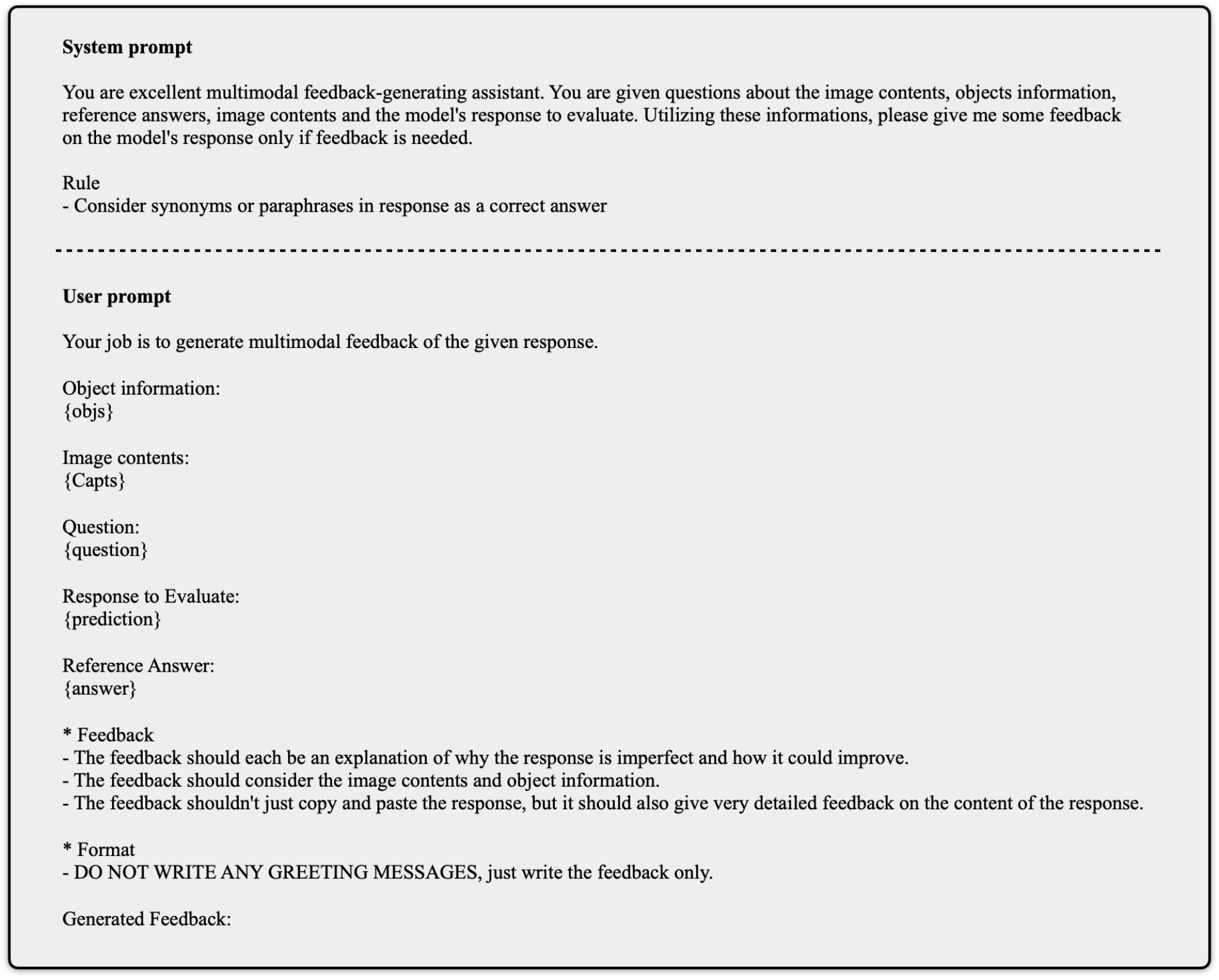}
\caption{\textbf{Prompt for generating multimodal feedback}}
\end{figure*}
\begin{figure*}[h]
\includegraphics[width=1.0\linewidth]{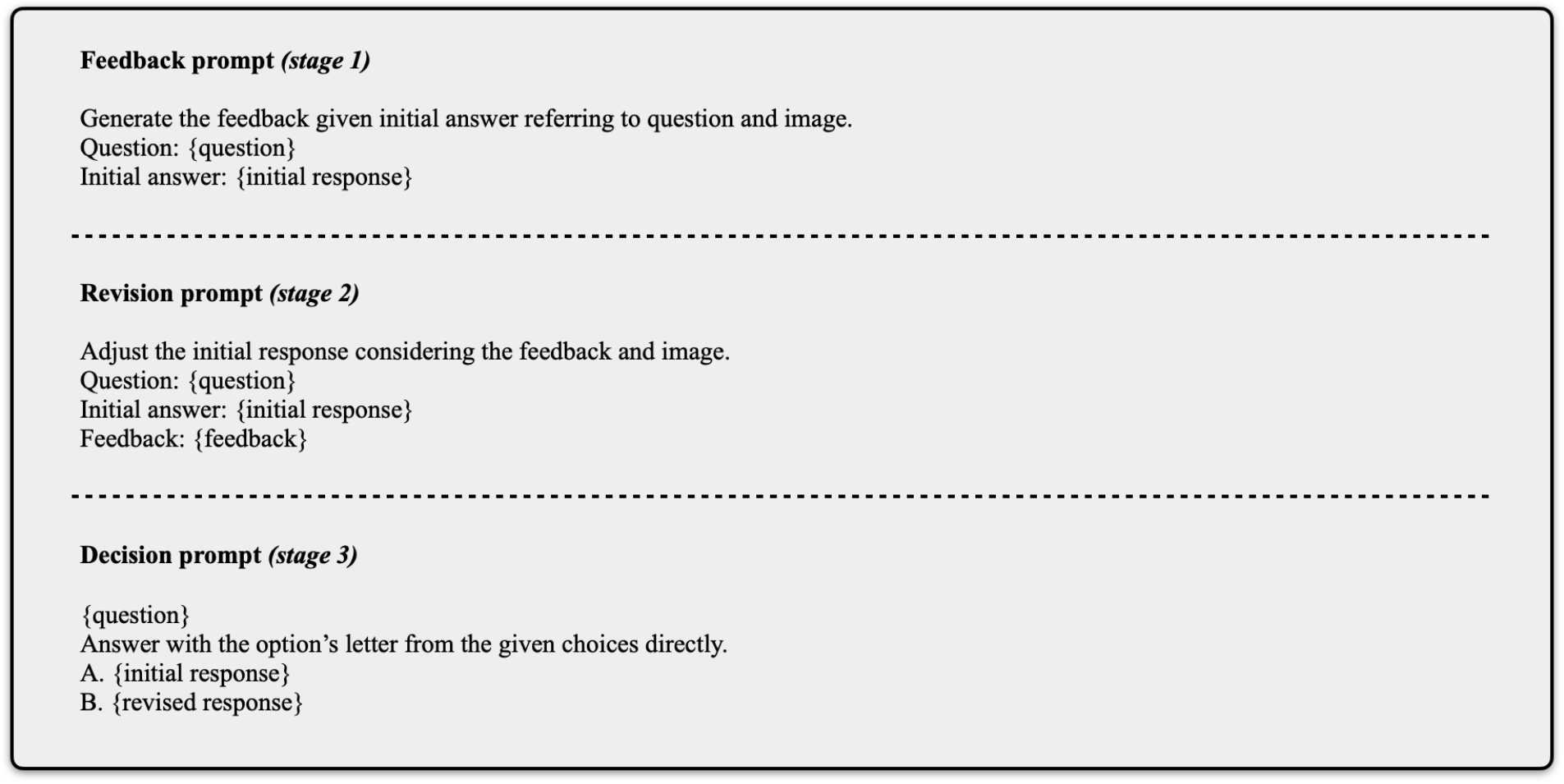}
\caption{\textbf{Prompts for inference at each stage}}
\end{figure*}

\end{document}